\documentclass[letterpaper, 10 pt, conference]{ieeeconf}  

\IEEEoverridecommandlockouts                              
\usepackage{hyperref}
\usepackage{type1cm}
\usepackage{makeidx}
\usepackage{amsmath}
\usepackage{graphicx}
\usepackage{subfig}
\usepackage{multicol}
\usepackage[bottom]{footmisc}
\usepackage{lipsum}
\usepackage{amsfonts}
\usepackage{color}
\usepackage[ruled,vlined,linesnumbered]{algorithm2e}
\usepackage{microtype}
\usepackage{authblk}
\usepackage{cite}

\usepackage{amsthm}
\let\labelindent\relax

\usepackage{enumitem}
\theoremstyle{plain}
\newtheorem{definition}{Definition}
\newtheorem*{problem*}{Problem}
\setlist[itemize]{align=parleft,left=0pt..1em}

\newcommand{\eg}{\textit{e.g.}}
\newcommand{\ie}{\textit{i.e.}}

\title{\LARGE \bf
Reactive Task and Motion Planning \\under Temporal Logic Specifications
}

\author{Shen Li\textsuperscript{\textdagger}*, 
Daehyung Park\textsuperscript{\textdagger\textdaggerdbl}*, 
Yoonchang Sung\textsuperscript{\textdagger}*,  
Julie A. Shah\textsuperscript{\textdagger}, 
and Nicholas Roy\textsuperscript{\textdagger} 
\thanks{*These authors contributed equally to this work. \textsuperscript{\textdagger}S. Li, D. Park, Y. Sung, J. Shah, and N. Roy are with CSAIL, Massachusetts Institute of Technology, USA. {\tt\small \{shenli,daehyung,yooncs8,julie\_a\_shah, nickroy\}@csail.mit.edu}. \textsuperscript{\textdaggerdbl}D. Park is also with the School of Computing, Korea Advanced Institute of Science and Technology, Korea. \\
This work was supported by Lockheed Martin Co. with the sponsor I.D. RPP2016-001 and the National Research Foundation of Korea(NRF) grant funded by the Korea government(MSIT) (No. 2021R1C1C1004368).
}
}

\begin{document}
\maketitle
\thispagestyle{empty}
\pagestyle{empty}

\begin{abstract}
We present a task-and-motion planning (TAMP) algorithm robust against a human operator's cooperative or adversarial interventions. Interventions often invalidate the current plan and require replanning on the fly. Replanning can be computationally expensive and often interrupts seamless task execution. We introduce a dynamically reconfigurable planning methodology with behavior tree-based control strategies toward reactive TAMP, which takes the advantage of previous plans and incremental graph search during temporal logic-based reactive synthesis. Our algorithm also shows efficient recovery functionalities that minimize the number of replanning steps. Finally, our algorithm produces a robust, efficient, and complete TAMP solution. Our experimental results show the algorithm results in superior manipulation performance in both simulated and real-world tasks.
\end{abstract}

\section{Introduction} \label{sec:1}

\begin{figure}[t]
\centering
    \includegraphics[trim={0cm 0cm 0cm 0cm},clip,width=0.32\textwidth]{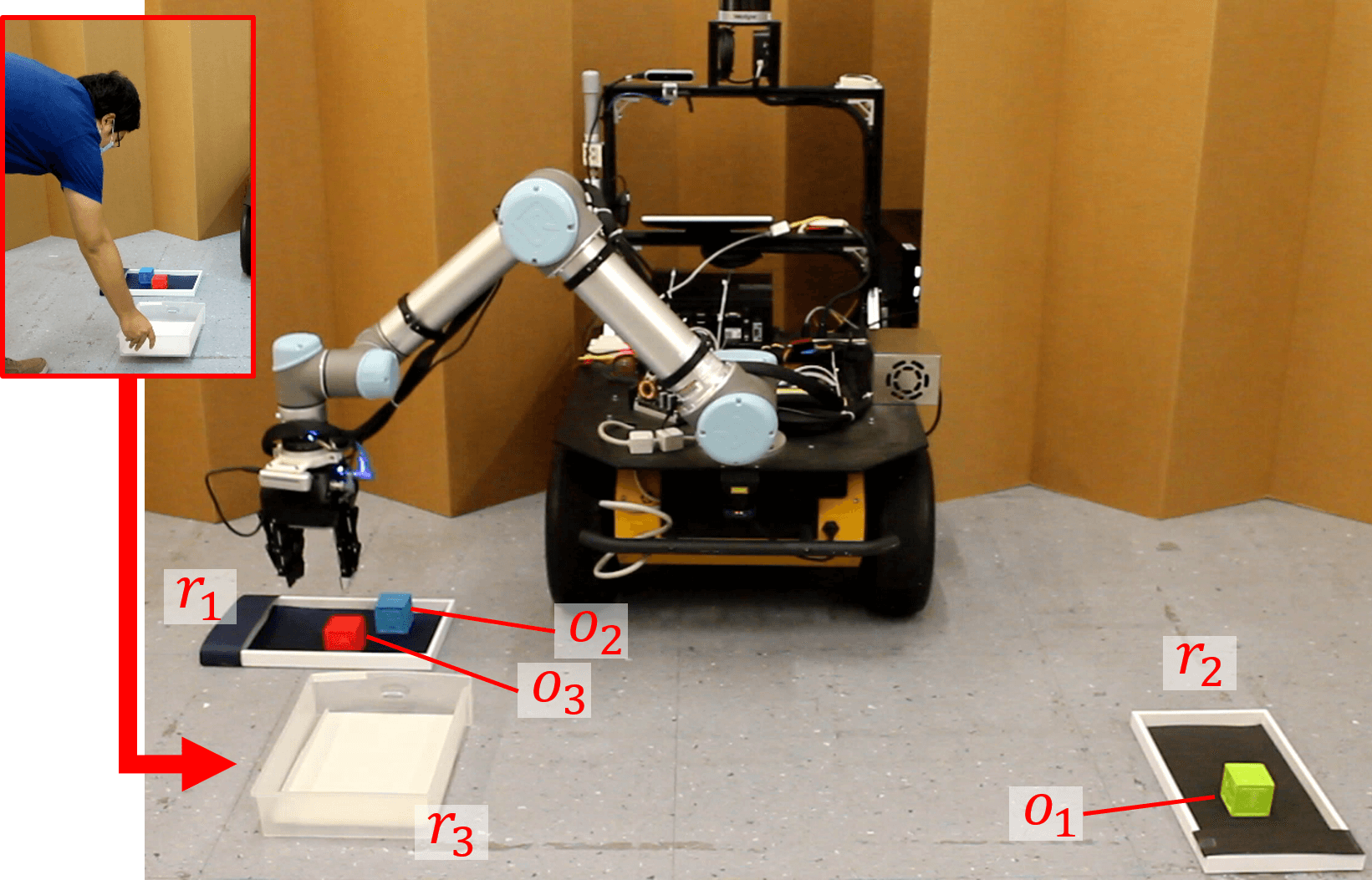}
\caption{\footnotesize{A capture of our ``3-block+tray" transfer task. Given a temporal logic-based goal specification \texttt{$\mathcal{F}\mathcal{G}(all\_obj\_in\_r_2)$}, our UR5 robot produces a complete task-and-motion plan to transfer objects from $r_1$ to $r_2$. When a human operator provides a movable tray $r_3$, our system reactively synthesizes a new plan that utilizes $r_3$ to produce efficient (\ie, minimum-distance cost) behaviors.} and also shows seamless execution using a behavior tree.}\label{fig: intro_exp}
\end{figure}

Consider the problem of collaborative operation between humans and robots~\cite{ajoudani2018progress}. A central challenge is responding in realtime to changes that might render the current robot task incomplete or inefficient. For example, we might like to ask a robot to ``pack up all the objects into the box'' as shown in Fig.~\ref{fig: intro_exp}. To achieve this complex goal, task-and-motion planning (TAMP) has been widely used for producing a sequence of actions \cite{dantam2016incremental,garrett2018sampling,vega2020asymptotically,srivastava2014combined}. During execution, a human partner might help the robot by packing up some of the objects or providing tools (\eg, a tray) to the robot. However, such changes may make the current plan invalid. The robot's plan might not have accounted for the new tool. Operating alongside human partners may require replanning-and-execution on the fly.


A popular replanning-and-execution approach is reactive synthesis that finds control strategies to react to environmental model changes including object position~\cite{he2019efficient, paxton2019representing, migimatsu2020object}, proposition~\cite{guo2013revising}, and uncertainty~\cite{livingston2012backtracking} while satisfying a given task specification~\cite{kress2009temporal,wolff2013efficient}. However, its reasoning process often requires the planner to construct a graph explicitly capturing feasible states (\ie, nodes) and edge connectivity between the states to find solutions. A wide range of model changes due to interventions leads to computationally expensive state-graph (re-)construction and search. Further, this graph reconstruction may lower execution efficiency, pausing ongoing motion until a new plan is reestablished. 

We propose a reactive TAMP method that hierarchically reacts to various levels of task interventions (\eg, the position changes of objects, addition/deletion of objects) by introducing complementary layers of planning and execution. The planning layer includes complete and efficient reactive synthesis using linear temporal logic (LTL) suitable to formally specify complex objectives and interventions~\cite{chinchali2012towards,he2015towards,plaku2016motion,verginis2018timed}. In detail, the method produces a sequence of action plans combining symbolic and (joint space) motion-cost based geometric searches over the LTL automata.
Given environmental changes, our first contribution is to show how to find a new plan with minimal search effort by incrementally expanding the search space and leveraging experience such as previously explored paths and corresponding costs. 
%

For the execution layer, we propose a behavior tree (BT) that conditionally activates sub-control policies (\ie, subtrees) to reactively follow action plans from the planning layer. BTs have been widely used to handle new or partially updated policies, as well as environmental changes~\cite{millington2009artificial,marzinotto2014towards,colledanchise2018behavior}. 
We use the modularity of BTs, which allows easier composition, addition, and deletion of policies on the fly, to partially update policies without pausing ongoing executions. Our second contribution is to show that the recovery strategies of conditional BTs \cite{giunchiglia2019conditional,paxton2019representing} can minimize expensive replanning steps with LTL. By relaxing the activation conditions of trees, we maximize the recovery range and minimize replanning requests given environment changes such as object relocation. 

In this work, we consider a robot that conducts a pick-and-place task of multiple objects where a human operator either cooperatively or adversarially could move the same objects. The robot can recognize the unexpected changes only by observing the location of objects. We assume that the robot does not suffer from partial and uncertain observations in this work. We do not explicitly model the human as another agent but rather consider as a part of the environment. We show our robot is able to adaptively change its plan on the fly to complete a given task. 
The key contributions include: 
\begin{itemize}
\item We present a robust TAMP method that hierarchically reacts to various environment changes or new temporal logic specifications. 
\item Our planning layer provides quickly replanned solutions based on past experience, online graph extension, and reconfigurable BT structures. 
\item Our execution layer allows seamless plan execution even with environment changes and computation delays.
\item We demonstrate that our method outperforms various baselines in terms of time efficiency and motion cost-effectiveness via physics-based simulations. We also conduct proof-of-concept experiments with a UR5 robot from Universal Robots. 
\end{itemize}

\section{Temporal-Logic Planning}
\label{sec:2}
We present a brief introduction to the LTL planning framework (more details in \cite{kress2018synthesis}) and show how we define variables used in the framework.

The planning problem studied in this work is to move a set of objects from initial locations to desired goal locations while overcoming environmental changes that might occur unpredictably. To enable temporal reasoning and guarantee task completion, we adopt temporal logic-based planning to synthesize a plan for the robot that achieves LTL task specifications, owing to its correct-by-design nature.

We model the robot and its workspace as a transition system ($\texttt{TS}$), which can be defined as tuples as follows:
\begin{definition}
(\textbf{Transition system})
$\texttt{TS}=(S,s_0,A,\delta_t,\Pi,$ $\mathcal{L})$,
where $S$ is a finite set of states ($s\in S$), $s_0$ is an initial state, $A$ is a set of actions, $\delta_t: S \times A \rightarrow S$ is a deterministic transition relation, $\Pi$ is a set of atomic propositions, and $\mathcal{L}: S \rightarrow 2^{\Pi}$ is a labeling function that maps each state to a subset of atomic propositions that hold true. 
\end{definition}

We compose an object $o_i$ with a region $r_j$ to denote a state $s$ in $\texttt{TS}$, \ie, $s=o_ir_j$, implying that the object $o_i$ is located in the region $r_j$. Let $O=\{o_1,...,o_n|o_i\in\mathbb{R}^3\}$ be the set of $n$ prehensile object states exist in the workspace of the robot, where $o_i$ is the 3D positional vector of the $i$-th object. The objects are placed in $m$ discrete and non-overlapping regions which we achieve by abstractly partitioning a continuous 3D workspace. We denote the set of $m$ region states by $R=\{r_1,...,r_m|r_j\in\mathbb{R}^3\}$. 
The robot can observe which objects belong to which region through its perception.
For multiple objects we represent a state $s$ by splitting pairs of an object $o_i$ and region $r_j$ by underscore, \eg, $s=o_1r_1\_o_2r_3$. The cardinality of $S$ is $|R|^{|O|}$. An element of the action set $A$ is represented by $\texttt{move}\_o_i\_r_j$ that is to pick an object $o_i$ from its current location and place $o_i$ in region $r_j$. An atomic proposition $\pi\in\Pi$ corresponds to a state $s\in S$ that is either true or false statements that must be specified for a given task. For example, $\pi=o_ir_j$ is true if an object $o_i$ is in a region $r_j$, or false otherwise. $\mathcal{L}$ maps a state $s \in S$ to a set of true atomic propositions $\mathcal{L}(s) \subseteq \Pi$.

An LTL formula $\varphi$ is a combination of Boolean operators and temporal operators, which is involved with a set of atomic propositions ($\subseteq \Pi$).
The formula $\varphi$ can express non-Markovian temporal properties, such as temporally-extended goals.
The syntax of the formula $\varphi$ over $\Pi$ is as follows:
\begin{equation}
\varphi=\pi\ |\ \neg\varphi\ |\  \varphi_1\vee\varphi_2\ |\ \mathcal{X}\varphi\ |\ \varphi_1\mathcal{U}\varphi_2,
\end{equation}
where we denote $\pi\in\Pi$, ``negation'' ($\neg$), ``disjunction'' ($\vee$), ``next'' ($\mathcal{X}$), and ``until'' ($\mathcal{U}$). We can additionally derive a Boolean operator ``conjunction'' ($\wedge$) and temporal operators ``always'' ($\mathcal{G}$) and ``eventually'' ($\mathcal{F}$) from the basic syntax. A detailed explanation can be found in~\cite{kress2018synthesis}.

A word is an infinite sequence of atomic propositions 
in $\Pi$ defined over the alphabet $2^\Pi$. There exists a union of inifinite words that satisfies a given formula $\varphi$. Alternatively, the nondeterministic B\"{u}chi automaton ($\texttt{BA}$) constructed according to~\cite{baier2008principles} is generally used to check the satisfiability of $\varphi$. 

\begin{definition}
(\textbf{B\"{u}chi automaton})
$\texttt{BA}=(Q_b,q_{b,0},\Sigma,\delta_b,$ $\mathcal{F}_b)$, where $Q_b$ is a finite set of states ($q_b\in Q_b$); $q_{b,0}$ is an initial state; $\Sigma=2^{\Pi}$ is an alphabet from the LTL formula; 
$\delta_b:Q_b\times\Sigma\rightarrow 2^{Q_b}$ is a non-deterministic transition relation; and $\mathcal{F}_b\subseteq Q_b$ is a set of accepting states.
\end{definition}

We then generate a product automaton ($\texttt{PA}$) by producting the $\texttt{TS}$ and $\texttt{BA}$ as follows.


\begin{definition}
(\textbf{Product automaton})
$\texttt{PA}=\texttt{TS}\times\texttt{BA}=(Q_p,q_{p,0},\Sigma,\delta_p,\mathcal{F}_p)$, where $Q_p=S\times Q_b$ whose element is $q_p=(s,q_b) \in Q_p$; $q_{p,0}=(s_0,q_{b,0})$; $\Sigma=2^{\Pi}$;
$\delta_p:Q_p\rightarrow Q_p$; and $\mathcal{F}_p\subseteq Q_p$.
Note that $\delta_p$ satisfies $(s^\prime,q_b^\prime)\in\delta_p(s,q_b)$ iff there exist $a \in A $ and $\delta_b$ such that $s^\prime=\delta_t(s,a)$ and $q_b^\prime\in\delta_b(q_b, \mathcal{L}(s))$.
\end{definition}

$\texttt{PA}$ becomes a search graph used for finding a plan. Finding a valid path that starts from an initial state $(s_0, q_{b,0})$ to one of accepting states $(s_F, q_{b,F})$ where $q_{b,F} \in \mathcal{F}_b$ implies computing a feasible plan $\xi = \{(s_i,q_{b,i})\}_{i = \{0,1,\dots,F\}}$ that achieves a given formula $\varphi$.

We study a scenario where unexpected human intervention affects the state of objects. A human operator might add new objects, remove existing objects, or move an object to a different region. This is equivalent to keeping the LTL specification $\varphi$ as it is while dynamically modifying the $\texttt{TS}$ and $\texttt{PA}$ to reflect the changes. Previous works also considered the same reactive approach as ours~\cite{livingston2012backtracking,guo2013revising}, or relaxed the LTL specification $\varphi$ instead while keeping the $\texttt{TS}$~\cite{guo2013reconfiguration,tumova2014maximally} for handling the changes. However, they mostly focused on navigation tasks.

The objective in this work is that the robot places all objects $O$ to desired goal regions among regions $R$. Although human intervention occurs unpredictably, the robot must be able to correct $\texttt{TS}$ and adjust its plan to accomplish a given task. Besides robustness, we would like the proposed replanning algorithm to additionally exhibit time efficiency and completeness.

\section{Dynamically Reconfigurable Planning}
\label{sec:3}

\begin{figure}[h]
\centering
\includegraphics[width=1.00\columnwidth]{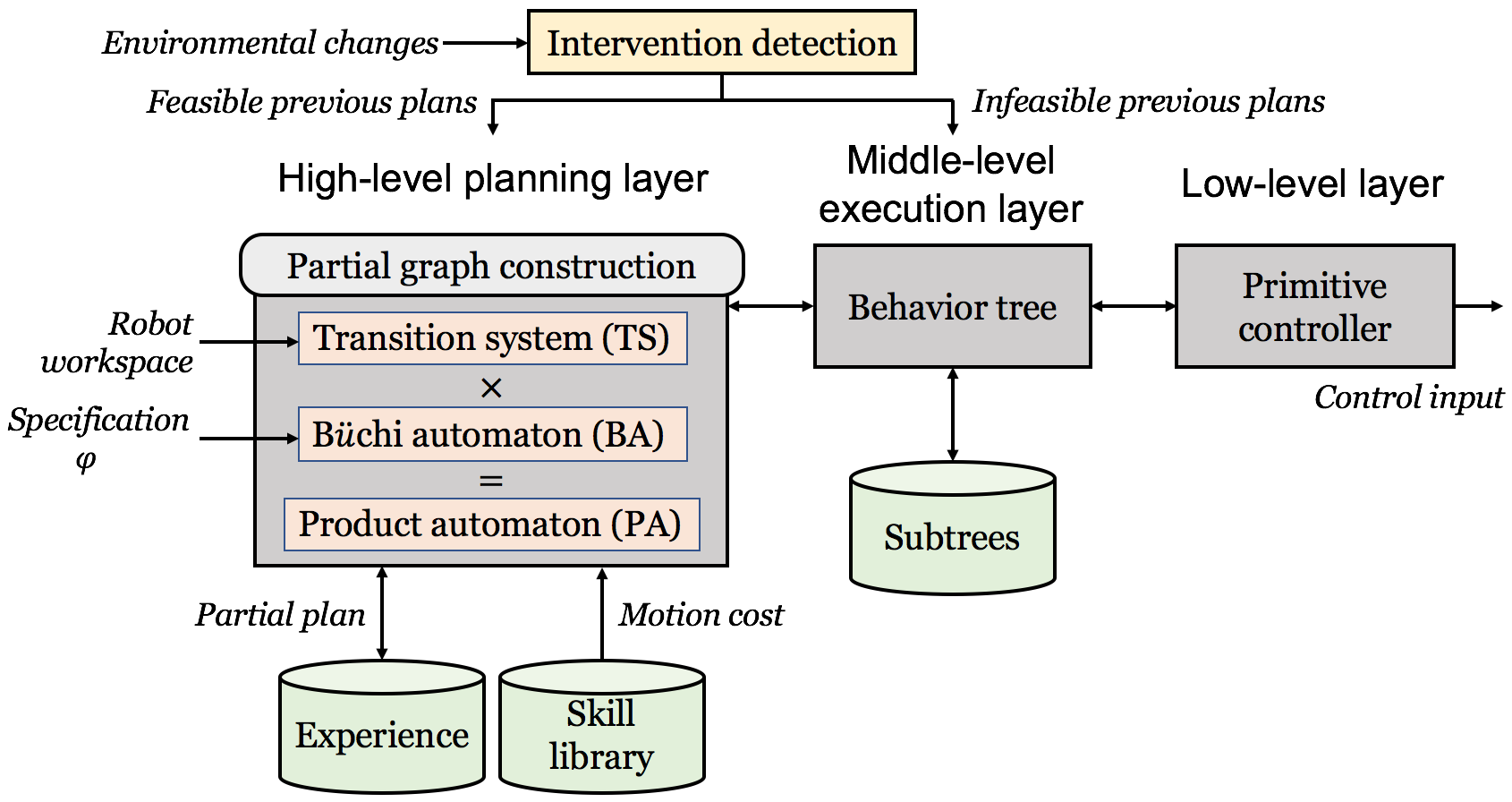}
\caption{\footnotesize{Overall hierarchical architecture. The high-level layer contains the $\texttt{TS}$, $\texttt{BA}$, and $\texttt{PA}$ that check the satisfiability of an LTL formula $\varphi$ as well as external databases, \ie, experience and skill library. The middle-level layer communicates between a high-level plan and a low-level controller via BT with a subtree database. The low-level layer outputs motor control commands to a robot. 
}}
\label{fig:flowchart}
\end{figure}

We address the proposed problem by directly finding a plan over the $\texttt{PA}$ that satisfies an LTL formula $\varphi$, \ie, placing all objects $O$ in desired goal regions $R$. We achieve this by a hierarchical architecture (Fig.~\ref{fig:flowchart}) that disentangles the high-level planning from low-level motor control commands. We handle unexpected changes by human intervention in upper layers of the architecture.

The entire proposed hierarchical architecture consists of three layers: (1) The high-level planning layer computes a high-level plan that satisfies LTL specifications; (2) The middle-level execution layer dynamically constructs BT to make high-level plans executable for low-level controller of the robot; and (3) The low-level layer outputs control inputs used in actuators. The inputs to the architecture are a robot model, information about workspace, and LTL specifications $\varphi$ and the ouput is low-level control inputs. Additional inputs are environmental changes that can be detected via perception.

Three desirable properties we aim at are robustness (against intervention), efficiency (in computing a plan), and completeness.
We achieve robustness efficiently by (1) handling intervention without involving the expensive task planner and only calling the planner when necessary; (2) reusing partial plans from the past; (3) computing a cost-effective plan by incorporating a motion cost computed by inverse kinematics; and (4) maintaining a partial graph of $\texttt{PA}$. We achieve completeness by employing an LTL-based planning.

In Section~\ref{subsec:hierarchical_system}, we present the capability of each component in our system and the pseudo-code of the framework.
In Section~\ref{subsec:planning_for_ltl_spec}, we introduce the search algorithm with motion cost to achieve a cost-effective solution.
In Section~\ref{subsec:partial_graph}, we present two ways of speeding up the search algorithms: reusing partial plans and maintaining a partial graph of \texttt{PA}.
In Section~\ref{subsec:bt}, we explain the BT reconstruction in detail.

\subsection{Robust Hierarchical System Design\label{subsec:hierarchical_system}}

In this subsection, we highlight which component of the hierarchical structure deals with which environmental changes. These components are in charge of different replanning queries that arrive at random time, including search graph construction and experience-based graph search in the high-level layer, and BT in the middle-level layer. In particular, we consider the following types of environmental changes: object relocation while previous plans are feasible or infeasible, and the addition and removal of objects and regions. Note again that the change in an LTL formula $\varphi$ is not considered.


The process of the system handling environmental changes is illustrated in Alg. \ref{alg:ltl_bt_sys}.
In line \ref{alg:ltl_bt_sys:relocated_if}, when the object is relocated to one of existing regions, there are cases where the previous plans can be reused if the plans have been reserved as \emph{subtrees} in the BT (details are explained in Section~\ref{subsec:bt}). 
In particular, if there are still executable subtrees (line \ref{alg:ltl_bt_sys:relocated_if_subtrees}), the expensive task planner will not be queried. For example, as shown in Fig. \ref{fig: intro_exp}, if the robot fails to pick up a block due to perception error, the BT will re-activate specific subtrees to attempt to grasp the block again.

Object relocation could also occur
where previous plans are no longer feasible. 
To handle this case, our system will use the same graph (skipping line \ref{alg:ltl_bt_sys:add_if_end}) and query a search algorithm to produce a new plan (line \ref{alg:ltl_bt_sys:replan}).
Particularly, the search algorithm (Section~\ref{subsec:planning_for_ltl_spec}) improves its planning efficiency by reusing the planning experiences computed and cached from the previous queries 
and maintaining a partial graph (Section~\ref{subsec:partial_graph}).
The BT will then receive the new plan and add/delete subtrees while keeping reusable subtrees for execution (line \ref{alg:ltl_bt_sys:BTreconfig}) (Section~\ref{subsec:bt}).

When the number of objects and that of regions change,
our system will reconstruct the \texttt{TS} and \texttt{PA} (line \ref{alg:ltl_bt_sys:add_if_end}) before querying the experience-based graph search for planning (line \ref{alg:ltl_bt_sys:replan}) and send the new plan to the BT for execution (line \ref{alg:ltl_bt_sys:BTreconfig}) (Section~\ref{subsec:bt}).

\subsection{Planning for LTL specifications\label{subsec:planning_for_ltl_spec}}
We use a weighted directed graph $G=(V,E,w)$ to represent the \texttt{PA} and deploy a graph search algorithm for planning. In particular, each $q_{p} \in Q_p$ corresponds to a node $v \in V$. For each $q_p$ and $q_p'$ where $q_p' \in \delta_p(q_p)$, there is an edge $e\in E$ between the nodes associated with $q_p$ and $q_p'$.



We incorporate the notion of motion cost into the $\texttt{PA}$ as a weight $w$ on an edge $e$ of $G$. We compute the motion cost as the incremental displacement of a robot's end effector, which can be obtained from a skill library assumed to be given. For example, when a manipulator approaches some location, the motion cost can be computed by inverse kinematics and saved as one of skills in the library. This allows for cost-effective plans with respect to a cumulative motion cost. 

To increase the planning efficiency, we propose two improvements over general graph search algorithms: (1) leveraging the computations from the past to reduce the computation in motion costs for a new replanning query; and (2) constructing $G$ on the fly to avoid the overhead of generating a graph from scratch whenever the world changes.

\begin{algorithm}[t]
\SetAlgoLined
\SetKwInput{KwInput}{Input}
\KwInput{\texttt{BA}, \texttt{TS}, $\Psi$}
$s$ $\leftarrow$ $s_0$; \tcp{Initial state}
$E$ $\leftarrow$ $\emptyset$; \tcp{Empty experience}
\While{Environmental changes}{
  $s_{new}$ $\leftarrow$ GetStateFromPerception()\;
  \If{Objects relocated}{\label{alg:ltl_bt_sys:relocated_if}
    $i$ $\leftarrow$ Find the subtree index $i$ of $(A^i, C_{pre}^i, C_{post}^i) \in T$ such that $C_{pre}^i \subset s_{new}$\;\label{alg:ltl_bt_sys:relocated_if_subtrees}
    \If{$i$ is found}{
      \tcp{Subtree $i$ can resolve this change by selecting action $A^i$ No replanning is needed}
      continue\;
    }
  }
  $s$ $\leftarrow$ $s_{new}$\;

  \If{Objects added/removed or regions added}{\label{alg:ltl_bt_sys:add_if}
    \tcp{Search graph reconstruction}
    \texttt{TS} $\leftarrow$ ReconstructTS(\texttt{TS}, $s$)\;\label{alg:ltl_bt_sys:add_if_end}
  }

  $q_{p,0}$ $\leftarrow$ $(s, q_{b,0})$; \tcp{Construct \texttt{PA} state}
  
  \tcp{Replan to find action tuples with partial graph construction, use cache experience (Sec. \ref{subsec:planning_for_ltl_spec}, \ref{subsec:partial_graph})}
  $P$, $E$ $\leftarrow$ LTLPlanner($q_{p,0}$, \texttt{BA}, \texttt{TS}, $E$)\;\label{alg:ltl_bt_sys:replan}
  \tcp{Update subtrees in BT (Sec. \ref{subsec:bt})}
  $\Psi$ $\leftarrow$ BTReconfiguration($P$, $\Psi$); \tcp{Alg. \ref{alg_bt_reconfig}}\label{alg:ltl_bt_sys:BTreconfig}
}
\caption{\footnotesize{\textbf{LTL-BT System}. Efficiently handling various types of interferences and updating the subtrees $\Psi$ in BT. 
}}\label{alg:ltl_bt_sys}
\end{algorithm}

\subsection{Search Acceleration}\label{subsec:partial_graph}


We attempt to speed up the search algorithm by reusing previous plans and partially constructing a search graph. When searching for a path in $G$, computing a motion cost on edge is costly as it requires solving an inverse kinematics problem. We cache the motion cost of edges newly explored so far as well as corresponding node names as a table data structure. If a new plan during replanning visits a node that has been previously visited, we look up the motion cost table to avoid inverse kinematics recomputation. Reusing previous plans expedites the search as environmental changes often affect only a small part of $G$.

Graph reconstruction is computationally expensive when the number of objects and that of regions vary.
The complexity of fully reconstructing the \texttt{PA} grows exponentially as the number of new objects increases.
The cardinality of $Q_p$ is $|S| \times |Q_b|$ where $|S| = |R|^{|O|}$.
To fully construct the graph $G$, the program needs to check for valid transitions for all pairs $\forall q_p, q_p' \in Q_p$, which results in $O(|Q_p|^2)=O(|R|^{2|O|}|Q_b|^2)$.
Our system constructs $G$ on the fly within the search (Alg.~\ref{alg:ltl_bt_sys} line \ref{alg:ltl_bt_sys:replan}) to reduce this overhead, rather than fully constructing it before the search.
With an admissible heuristic, our system can avoid expanding all $v \in V$ and avoid constructing the entire $G$ while finding the same path when heuristics are not used.
In particular, within the search algorithm (line \ref{alg:ltl_bt_sys:replan}), when a node associated with $q_p$ is chosen to be expanded, the system will query $\delta_p$ to produce the list of next states $\{q_p'\}=\delta_p(q_p)$ and add them to $V$ and $E$ accordingly.
In the worst case, if the heuristic is not effective at all, all nodes in $V$ will be expanded, and $G$ will be fully constructed.
We call this approach \textit{partial graph construction}. We show the efficiency of partial graph construction against the approach of fully constructing $G$ every time before the search (\textit{full graph construction}) in Section~\ref{sec:3}.

%

\subsection{Behavior Tree \& Reconfiguration}~\label{subsec:bt}
\begin{figure}[h]
	\centering
    \subfloat[With state-based conditions]{
    \includegraphics[trim={0cm 0cm 0cm 0cm},clip,width=0.23\textwidth]{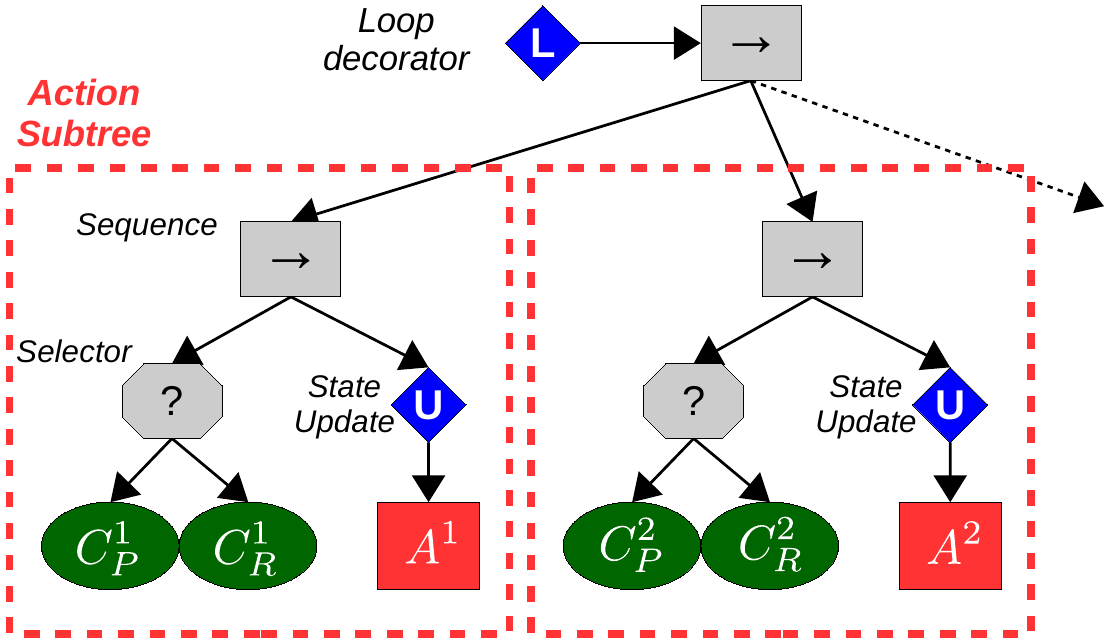}
    \label{fig_subtree1}
    }
    \hspace{-2mm}
    \subfloat[With action-based conditions]{
    \includegraphics[trim={0cm 0cm 0cm 0cm},clip,width=0.23\textwidth]{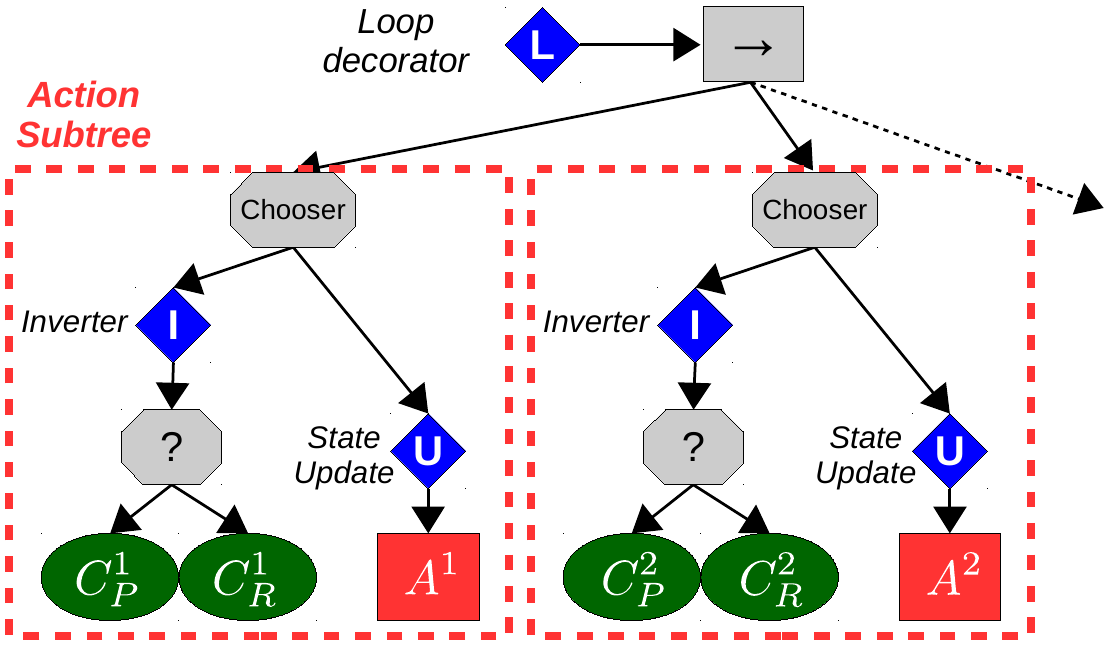}
    \label{fig_subtree2}
    }
\caption{\footnotesize{Subtree structures. Chooser is a custom selector that does not interrupt the low priority of a child node from the higher priority of nodes.}}\label{fig: subtree}
\end{figure}

The BT is a compositional decision-making system that can group a set of behaviors, execute between groups at a fixed frequency, and change the structure on the fly. The execution starts from the root of the BT. It then gradually  traverses the execution permission (the tick) from the root to child nodes. Once a node receives the tick, the node execution returns: 1) SUCCESS if the node execution was successfully completed, 2) RUNNING if the execution is not completed yet, 3) FAILURE if the execution was failed, and 4) INVALID otherwise. Fig.~\ref{fig: subtree} shows the execution flow given a sequence of action plans, where $\rightarrow$ and $?$ represent \textit{sequence} and \textit{selector} behaviors, respectively (see details in \cite{pytrees}).

There are previous works~\cite{tumova2014maximally,colledanchise2017synthesis,lan2019autonomous} adopting both the LTL and BT for synthesizing controls from high-level plans. However, their methods cannot deal with environmental changes studied in this work.

We now describe how to build the internal structure of the BT given a high-level plan and reconfigure it on the fly.

\subsubsection{Automatic tree construction}
We convert a sequence of action plans $P$ to a sequence of subtrees $\Psi$, which can be attached to the task root of a BT. Similar to the planning domain description language (PDDL~\cite{fox2003pddl2}), we use preconditions and effects to achieve complex sequences of behaviors similar to \cite{giunchiglia2019conditional, lan2019autonomous}. To build the tree, we parse a plan $P$ into a sequence of action tuples $T=\{(A^1, C^1_{pre}, C^1_{post}), (A^2, C^2_{pre}, C^2_{post}), ... \}$, where $A^i$, $C^i_{pre}$, and $C^i_{post}$ represent an action, a pre-condition, and a post-condition, respectively. For example, given two consecutive states $(o_1r_1\_o_2r_1, o_1r_2\_o_2r_1)$, the action is $o_1$'s move from $r_1$ to $r_2$, where $C_{pre}=\{o_1r_1,o_2r_1\}$ and $C_{post}=\{o_1r_2,o_2r_1\}$. 

We then convert the tuple to a \textit{state condition}-based subtree as shown in Fig.~\ref{fig_subtree1}, where $C_R$ represents a set of running conditions (\eg, $o_1r_{\text{hand}}\_o_2r_1$) from \cite{paxton2019representing} and the \textit{State Update} decorator forces the current state to be updated after the completion of an action. Note that $C^i_{post}$ equals to $C^{i+1}_{pre}$ in this sequential plan so that we can ignore post conditions. This structure can also re-execute the previous actions if the current state satisfies the precondition. However, the state-based condition may be too conservative in that a full state specification cannot ignore unrelated environmental changes. 

Alternatively, we introduce a relaxed version of subtree constructor, the \textit{action condition}-based subtree (see Fig.~\ref{fig_subtree2}), that is designed to prevent unnecessary replanning given unrelated or minor environmental changes such as object additions, deletions, and repositions. The subtree also allows more recovery behaviors. In particular, the precondition node checks soft conditions including the location of the target object, object source, and object destination (\eg, $C_{\text{pre}}=o_1r_1$).
We also use a customized \textit{selector}, called \textit{chooser}~\cite{pytrees}, as a root of each subtree that is to prevent interruption from higher priority of sibling subtrees.
This enables BT to complete an ongoing action.

\subsubsection{Online reconfiguration}
We introduce how we can reconfigure a BT without interfering with running tasks (\ie, running subtrees) on the fly. Given environmental changes, the planner will produce a new plan $P'$, which may not be completely different from the previous plan $P$ so it is not necessary to reconstruct a sequence of subtrees $S$ from scratch. Alg.~\ref{alg_bt_reconfig} shows how we can find sharable subtrees between $P$ and $P'$ from the back. If there is a new action in $P'$, we add a new subtree to $S$. Otherwise, we remove the unmatched subtrees. Note that if there are any running and non-interruptible subtrees, our framework postpones the change of the subtree until the return of the subtree is changed.  


\begin{algorithm}[t]
\SetAlgoLined
\DontPrintSemicolon
\SetKwInput{KwOutput}{Output}
\KwOutput{$\Psi$}
 $P \leftarrow (P^1, ... , P^n)$ \tcp*{A seq. of action tuples}
 $\Psi \leftarrow (\Psi^1, ... , \Psi^m)$ \tcp*{A seq. of subtrees}
 \For{$i\gets1$ \KwTo n}{
  \uIf{$P_A^{-i}=\Psi_A^{-i}  \:\&\: |\Psi| \geq i $}{
   $\Psi_C^{-i} \leftarrow P^{-i}_C$ \tcp*{Update conditions}
  }
  \uElseIf{$ P_A^{-i} \neq \exists \Psi_A^{:-i}  $}{
  $j \leftarrow \text{find the index of }P_A^{-i-1}\text{ at }\Psi_A^{1:-i} $\;
  \text{Remove subtrees at $\Psi^{j:-i}$}\;
  }
  \Else{
  \text{Add a subtree for $P^{-i}$ at $\Psi^{-i}$}\;
  }
 }
 
 \If{$|\Psi|\neq n$}{
\text{Remove $\forall s \in \Psi^{1:|\Psi|-n}$}\;
}
\caption{\footnotesize{\textbf{BT Reconfiguration}. $P_A^i$ and $\Psi_A^i$ represent the action name of the $i$-th action tuple and subtree, respectively. Likewise, $P$ and $A$ with the subscript $C$ represent their preconditions. The superscript $-i$ accesses the $i$th element from last.}}\label{alg_bt_reconfig}
\end{algorithm}

%

\section{Experimental Setup}

\subsection{Statistical Evaluation}
We evaluated our method with baselines on two environment settings in a physics-based simulator, GAZEBO (Ver.~7)~\cite{koenig2004design}. The manipulation settings were composed of colored blocks, trays, and/or baskets (see Fig.~\ref{fig: demo}). Given an LTL specification, \texttt{$\varphi=\mathcal{F}\mathcal{G}(all\_obj\_in\_r_2)$}, our method generated a plan for transferring all blocks from $r_1$ to $r_2$. 
Note that other LTL specifications can also be handled in our framework. 
Per each method and setting, we created 30 test scenarios by randomly placing objects in the beginning and then introduced an additional environmental change, such as repositioning, adding, or deleting a randomly selected object at a random time. Descriptions of each scenario and setting are as follow:
1) In the \textit{3-block} scenario, as shown in Fig. \ref{fig: demo}(a), there are three blocks initially located at $r_1$. The desired plan is to move each block one after another from $r_1$ to $r_2$.
2) In the \textit{3-block + tray} scenario, as shown in Fig. \ref{fig: demo}(b), there are three blocks initially located at $r_1$ and a movable tray $r_3$ near $r_1$. The optimal plan is to transfer all the blocks to $r_3$ first, move $r_3$ near $r_2$, and then relocate blocks from $r_3$ to $r_2$. By leveraging $r_3$, the robot can reduce the total motion costs of transferring blocks from $r_1$ to $r_2$.

\subsection{Demonstration Setup}
We demonstrated the proposed method on two manipulation scenarios of a real UR5 robot (see Fig.~\ref{fig: demo}).
The robot observed objects using an Intel D435 RGB-D camera mounted on the wrist. It first recognized object segments by using \textit{Detectron2}~\cite{wu2019detectron2} with \textit{Mask R-CNN}~\cite{he2017mask} from Facebook, where we trained Detectron2 images collected from similar settings after labeling it via COCO Annotator~\cite{cocoannotator}. The segments were then used for estimating the 3D pose and grasping point of each object from the depth image. Here, we assumed there are unique colors of objects to assign unique ID. An operator randomly relocated, added, or removed objects including a tray during the experiment.


\section{Results}
\label{sec:3}

\begin{figure}[t]
\centering
\includegraphics[trim={0cm 0cm 0cm 0cm},clip,width=0.45\textwidth]{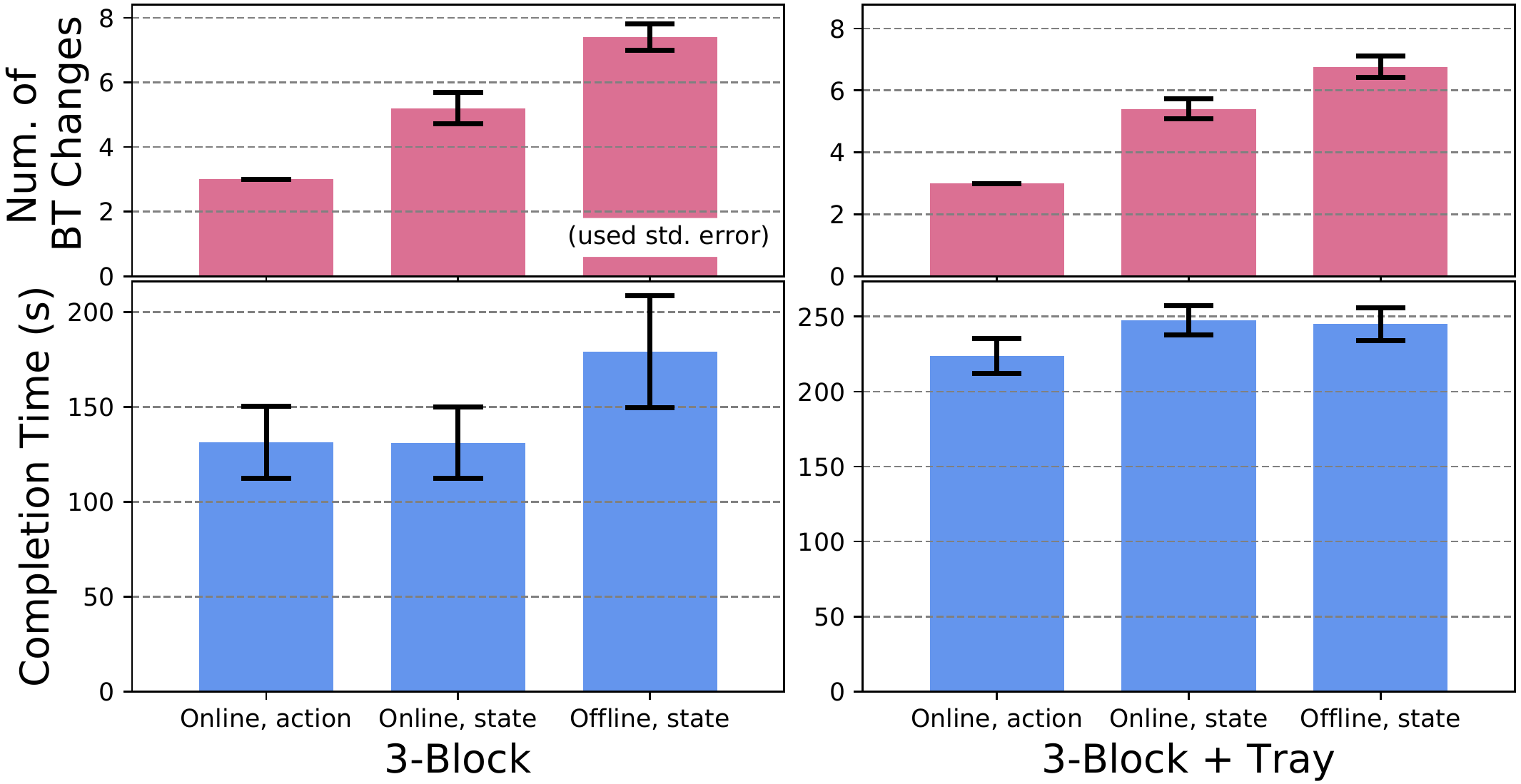}
\caption{\footnotesize{Comparison of on/offline BT with \textit{action} and \textit{state} conditions.}}\label{fig:bt_reconfig}
\end{figure}

\begin{figure*}[ht]
	\centering
    \subfloat[A ``3-block" transfer task. An operator added and then repositioned objects during the task.]{
    \includegraphics[trim={0cm 0cm 0cm 0cm},clip,width=0.90\textwidth]{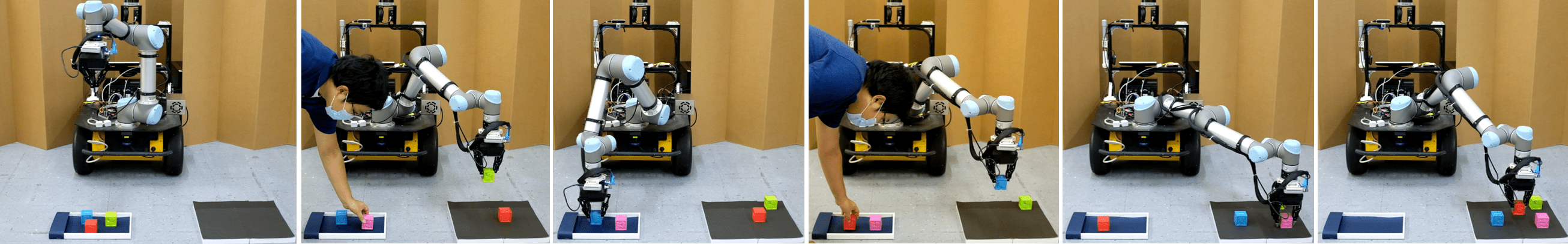}
    \label{fig: demo_a}}  
    \\    
    \subfloat[A ``3-block + tray" task. An operator provided a movable tray during the task.]{
    \includegraphics[trim={0cm 0cm 0cm 0cm},clip,width=0.98\textwidth]{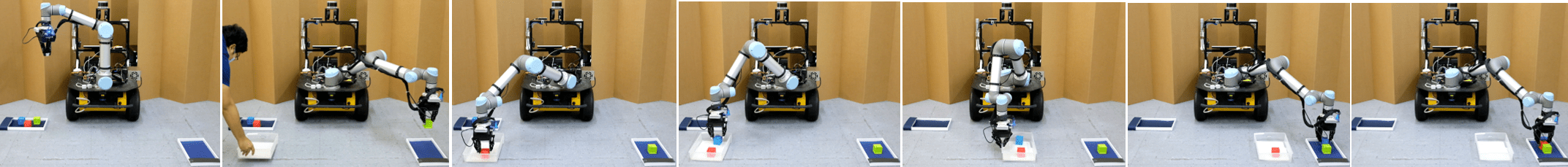} \label{fig: demo_b}}
\caption{\footnotesize{Demonstrations on the \textit{3-block} and \textit{3-block+tray} scenarios, where a UR5 robot is commended to place all objects on the right place. A human operator artificially generated environmental changes during the object transfer tasks. The robot with the proposed TAMP method successfully completed the task by producing a reactive and cost-effective solution.
Please see the attached video: \url{https://youtu.be/lPpMVfBzZH0}.
}}\label{fig: demo}
\end{figure*}

We benchmarked our method against a set of baselines and demonstrated its robustness and efficiency via simulations.
To show the performance of the execution layer, we compared three variations of the BT: \textit{online} reconfiguration with \textit{action condition}, \textit{online} reconfiguration with \textit{state condition}, and \textit{offline} reconstruction with \textit{action condition} over two scenarios. Note that the offline reconstruction stops a running node to remove all subtrees. It then constructs a new tree from scratch in an idle mode. 
%
Fig.~\ref{fig:bt_reconfig} shows the \textit{online} reconfiguration with \textit{action condition} resulted in a smaller number of BT changes and took less completion time compared with that of \textit{state condition}. Due to the short replanning time of A* with experience, the online method with \textit{action} and \textit{state} conditions resulted in a similar completion time in the 3-Block scenario.
On the other hand, the \textit{offline} reconstruction incurred longer average completion time than \textit{online} BT.
%
Overall, the results demonstrate that the \textit{online} BT reconfiguration with \textit{action condition} is the most efficient among the variations.

We benchmarked three search algorithms, A$^*$ with experience, A$^*$, and Dijkstra, in the 3-block + tray scenario, each with three types of environmental changes (repositioning, adding, or deleting a block).
We split all planning processes in each trial into (1) initial planning in the very beginning and (2) replannings, used for intervention handling.
Table~\ref{table:search_gazebo} highlights the robustness and efficiency of our hierarchical system.
The column ``$\#$ replan'' presents that both means and medians of the number of replannings for object repositions are lower than those for additions and deletions across all algorithms. This result demonstrates that the system's efficiency in handling repositions by using the BT alone without the expensive replanning unless necessary.
The mean of the number of replannings for object reposition is non-zero because sometimes it is necessary to replan.
For example, if $o_i$ is repositioned to $r_1$ when $r_3$ is near $r_2$, since the state $o_ir_1\_r_3r_2$ is not in the optimal initial plan, no subtrees in BT can handle this case alone wihtout replanning.
Table~\ref{table:search_gazebo} highlights the efficiency of the experience-based search. A$^*$ with experience leverages the similarity between the initial plan and the plans for intervention handling. In Table~\ref{table:search_gazebo}, A$^*$ with experience has a significantly lower total replanning time than A$^*$ and Dijkstra among all types of environmental changes. The initial planning time is almost the same across three algorithms due to a lack of experience in the very beginning.
To demonstrate the overall efficiency, we report the completion time of each trial. Among all types of environmental changes, A$^*$ with experience achieves significantly shorter completion time due to faster replanning.
%
The success rate is always $100\%$ in Table~\ref{table:search_gazebo} as our system guarantees task completion.

\begin{table}[h]
    \centering
    \resizebox{\columnwidth}{!}{%
    \begin{tabular}{l|l|l|l|l|l}
    \hline
        3-block + tray & Success & Init plan time (s) & Total replan time (s) & $\#$ replan & Comple time (s)\\ \hline
        Rel, A$^*$, exp & $30/30$ & $68.39 \pm 1.00$ & $\mathbf{0.69 \pm 0.43}$ & $0.80$, $0$ & $\mathbf{466.50 \pm 34.80}$\\
        Rel, A$^*$ & $30/30$ & $67.19 \pm 0.04$ & $33.50 \pm 15.26$ & $0.63$, $0$ & $530.17 \pm 19.04$\\
        Rel, Dijkstra & $30/30$ & $67.36 \pm 0.04$ & $30.40 \pm 13.06$ & $0.47$, $0$ & $561.57 \pm 27.29$\\
        \hline
        Rem, A$^*$, exp & $30/30$ & $67.62 \pm 0.22$ & $\mathbf{0.10 \pm 0.02}$ & $1.93$, $1$ & $\mathbf{394.48 \pm 13.14}$\\
        Rem, A$^*$ & $30/30$ & $67.34 \pm 0.07$ & $52.57 \pm 8.18$ & $1.30$, $1$ & $536.25 \pm 31.49$\\
        Rem, Dijkstra & $30/30$ & $67.57 \pm 0.08$ & $60.53 \pm 9.16$ & $1.40$, $1$ & $478.43 \pm 17.54$\\
        \hline
        Add, A$^*$, exp & $30/30$ & $67.78 \pm 0.32$ & $\mathbf{20.61 \pm 0.86}$ & $1.37$, $1$ & $\mathbf{647.00 \pm 24.67}$\\
        Add, A$^*$ & $30/30$ & $69.16 \pm 1.29$ & $89.60 \pm 8.73$ & $1.13$, $1$ & $734.74 \pm 33.32$\\
        Add, Dijkstra & $30/30$ & $67.60 \pm 0.10$ & $116.16 \pm 16.35$ & $1.33$, $1$ & $720.81 \pm 25.03$\\
        \hline
    \end{tabular}
    }
    \caption{\footnotesize{Results of statistical evaluation regarding 3 search algorithms (A$^*$ with planning experiences, A$^*$, and Dijkstra), with partial graph construction, in handling 3 types of environmental changes (relocating 1 object, removing 1 object, and adding 1 object). The metrics used here are number of successes/number of trials, initial planning time ($95\%$ confidence interval (CI)), total replanning time ($95\%$ CI), number of replannings (mean, median), and completion time ($95\%$ CI).}}
    \label{table:search_gazebo}
\end{table}

\begin{figure}[h]
\centering
\includegraphics[trim={0cm 0cm 0cm 0cm},clip,width=0.32\textwidth]{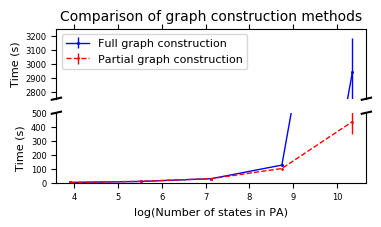}
\caption{\footnotesize{Comparison of time taken for constructing the \texttt{PA} and replanning between the full graph construction and the partial graph construction.
}}\label{fig:partial}
\end{figure}

We validated the efficiency of the partial graph construction over the full graph construction through simulations. We measured the time taken for both constructing the $\texttt{PA}$ and replanning when an object was added as environmental changes for efficiency comparison. We fixed the number of regions to be five while varying the number of objects from two to six, resulting in the number of states in the $\texttt{PA}$ to vary from $50$ to $31,250$. We randomly generated locations of regions ten times. Fig.~\ref{fig:partial} shows the mean and the standard deviation of a total time measured over ten random environments for both graph construction methods. The full graph construction performed comparably with the partial graph construction up to $1,250$ states but started degrading exponentially afterward.

We finally demonstrated the proposed TAMP method with a real UR5 robot. Fig.~\ref{fig: demo} shows captures of two representative intervention scenarios. Given the same LTL specification, our robot successfully initiated a complete task and motion plan to transfer all objects to another storage by visually recognizing the environment. When an operator added and relocated objects, the online BT enabled the robot to continue seamless task execution while replanning the new setup. The proposed graph construction and experience-based planning enabled the robot quickly to transition to a new plan. Given a new transfer tool (see Fig.~\ref{fig: demo_b}), our system could also show cost-effective behaviors by utilizing the tray as we humans do. These demonstrations show our framework enables the robot to achieve robust and efficient high-level planning and middle-level execution in human-robot operation scenarios.

\section{Discussion \& Conclusion}
We introduce a hierachically reactive TAMP method, which generates a complete and semi-optimal task plan by using LTL and approximated motion costs. Our framework provides an efficient replanning approach leveraging experience-based graph search and node modification algorithms. In addition, to minimize the replanning and also maximize the recovery control capability, we combine it with a dynamically reconfigurable BT with the \textit{action condition} subtree structure. Our statistical evaluation shows the superial efficiency compared to other baseline methods. We also demonstrate the robustness in real world with a UR5 robot. 

\bibliographystyle{ieeetr}
\bibliography{references}

\begin{thebibliography}{10}

\bibitem{ajoudani2018progress}
A.~Ajoudani, A.~M. Zanchettin, S.~Ivaldi, A.~Albu-Sch{\"a}ffer, K.~Kosuge, and
  O.~Khatib, ``Progress and prospects of the human--robot collaboration,'' {\em
  Autonomous Robots}, vol.~42, no.~5, pp.~957--975, 2018.

\bibitem{dantam2016incremental}
N.~T. Dantam, Z.~K. Kingston, S.~Chaudhuri, and L.~E. Kavraki, ``Incremental
  task and motion planning: A constraint-based approach.,'' in {\em Proceedings
  of Robotics: Science and Systems (RSS)}, vol.~12, p.~00052, Ann Arbor, MI,
  USA, 2016.

\bibitem{garrett2018sampling}
C.~R. Garrett, T.~Lozano-P{\'e}rez, and L.~P. Kaelbling, ``Sampling-based
  methods for factored task and motion planning,'' {\em International Journal
  of Robotics Research}, vol.~37, no.~13-14, pp.~1796--1825, 2018.

\bibitem{vega2020asymptotically}
W.~Vega-Brown and N.~Roy, ``Asymptotically optimal planning under
  piecewise-analytic constraints,'' in {\em Algorithmic Foundations of Robotics
  XII}, pp.~528--543, Springer, 2020.

\bibitem{srivastava2014combined}
S.~Srivastava, E.~Fang, L.~Riano, R.~Chitnis, S.~Russell, and P.~Abbeel,
  ``Combined task and motion planning through an extensible planner-independent
  interface layer,'' in {\em Proceedings of the IEEE International Conference
  on Robotics and Automation (ICRA)}, pp.~639--646, IEEE, 2014.

\bibitem{he2019efficient}
K.~He, A.~M. Wells, L.~E. Kavraki, and M.~Y. Vardi, ``Efficient symbolic
  reactive synthesis for finite-horizon tasks,'' in {\em Proceedings of the
  IEEE International Conference on Robotics and Automation (ICRA)},
  pp.~8993--8999, IEEE, 2019.

\bibitem{paxton2019representing}
C.~Paxton, N.~Ratliff, C.~Eppner, and D.~Fox, ``Representing robot task plans
  as robust logical-dynamical systems,'' in {\em Proceedings of the IEEE/RSJ
  International Conference on Intelligent Robots and Systems (IROS)},
  pp.~5588--5595, IEEE, 2019.

\bibitem{migimatsu2020object}
T.~Migimatsu and J.~Bohg, ``Object-centric task and motion planning in dynamic
  environments,'' {\em IEEE Robotics and Automation Letters}, vol.~5, no.~2,
  pp.~844--851, 2020.

\bibitem{guo2013revising}
M.~Guo, K.~H. Johansson, and D.~V. Dimarogonas, ``Revising motion planning
  under linear temporal logic specifications in partially known workspaces,''
  in {\em Proceedings of the IEEE International Conference on Robotics and
  Automation (ICRA)}, pp.~5025--5032, IEEE, 2013.

\bibitem{livingston2012backtracking}
S.~C. Livingston, R.~M. Murray, and J.~W. Burdick, ``Backtracking temporal
  logic synthesis for uncertain environments,'' in {\em Proceedings of the IEEE
  International Conference on Robotics and Automation (ICRA)}, pp.~5163--5170,
  IEEE, 2012.

\bibitem{kress2009temporal}
H.~Kress-Gazit, G.~E. Fainekos, and G.~J. Pappas, ``Temporal-logic-based
  reactive mission and motion planning,'' {\em Transactions on Robotics},
  vol.~25, no.~6, pp.~1370--1381, 2009.

\bibitem{wolff2013efficient}
E.~M. Wolff, U.~Topcu, and R.~M. Murray, ``Efficient reactive controller
  synthesis for a fragment of linear temporal logic,'' in {\em Proceedings of
  the IEEE International Conference on Robotics and Automation (ICRA)},
  pp.~5033--5040, IEEE, 2013.

\bibitem{chinchali2012towards}
S.~Chinchali, S.~C. Livingston, U.~Topcu, J.~W. Burdick, and R.~M. Murray,
  ``Towards formal synthesis of reactive controllers for dexterous robotic
  manipulation,'' in {\em Proceedings of the IEEE International Conference on
  Robotics and Automation (ICRA)}, pp.~5183--5189, IEEE, 2012.

\bibitem{he2015towards}
K.~He, M.~Lahijanian, L.~E. Kavraki, and M.~Y. Vardi, ``Towards manipulation
  planning with temporal logic specifications,'' in {\em Proceedings of the
  IEEE International Conference on Robotics and Automation (ICRA)},
  pp.~346--352, IEEE, 2015.

\bibitem{plaku2016motion}
E.~Plaku and S.~Karaman, ``Motion planning with temporal-logic specifications:
  Progress and challenges,'' {\em AI communications}, vol.~29, no.~1,
  pp.~151--162, 2016.

\bibitem{verginis2018timed}
C.~K. Verginis and D.~V. Dimarogonas, ``Timed abstractions for distributed
  cooperative manipulation,'' {\em Autonomous Robots}, vol.~42, no.~4,
  pp.~781--799, 2018.

\bibitem{millington2009artificial}
I.~Millington and J.~Funge, {\em Artificial intelligence for games}.
\newblock CRC Press, 2009.

\bibitem{marzinotto2014towards}
A.~Marzinotto, M.~Colledanchise, C.~Smith, and P.~{\"O}gren, ``Towards a
  unified behavior trees framework for robot control,'' in {\em Proceedings of
  the IEEE International Conference on Robotics and Automation (ICRA)},
  pp.~5420--5427, IEEE, 2014.

\bibitem{colledanchise2018behavior}
M.~Colledanchise and P.~{\"O}gren, {\em Behavior trees in robotics and AI: An
  introduction}.
\newblock CRC Press, 2018.

\bibitem{giunchiglia2019conditional}
E.~Giunchiglia, M.~Colledanchise, L.~Natale, and A.~Tacchella, ``Conditional
  behavior trees: Definition, executability, and applications,'' in {\em
  Proceedings of the IEEE International Conference on Systems, Man and
  Cybernetics (SMC)}, pp.~1899--1906, IEEE, 2019.

\bibitem{kress2018synthesis}
H.~Kress-Gazit, M.~Lahijanian, and V.~Raman, ``Synthesis for robots: Guarantees
  and feedback for robot behavior,'' {\em Annual Review of Control, Robotics,
  and Autonomous Systems}, 2018.

\bibitem{baier2008principles}
C.~Baier and J.-P. Katoen, {\em Principles of model checking}.
\newblock MIT press, 2008.

\bibitem{guo2013reconfiguration}
M.~Guo and D.~V. Dimarogonas, ``Reconfiguration in motion planning of
  single-and multi-agent systems under infeasible local ltl specifications,''
  in {\em Proceedings of the IEEE Conference on Decision and Control},
  pp.~2758--2763, IEEE, 2013.

\bibitem{tumova2014maximally}
J.~Tumova, A.~Marzinotto, D.~V. Dimarogonas, and D.~Kragic, ``Maximally
  satisfying ltl action planning,'' in {\em Proceedings of the IEEE/RSJ
  International Conference on Intelligent Robots and Systems (IROS)},
  pp.~1503--1510, IEEE, 2014.

\bibitem{pytrees}
``Py trees.'' http://py-trees.readthedocs.io.
\newblock Accessed: 2020-07-10.

\bibitem{colledanchise2017synthesis}
M.~Colledanchise, R.~M. Murray, and P.~{\"O}gren, ``Synthesis of
  correct-by-construction behavior trees,'' in {\em Proceedings of the IEEE/RSJ
  International Conference on Intelligent Robots and Systems (IROS)},
  pp.~6039--6046, IEEE, 2017.

\bibitem{lan2019autonomous}
M.~Lan, S.~Lai, T.~H. Lee, and B.~M. Chen, ``Autonomous task planning and
  acting for micro aerial vehicles,'' in {\em Proceedings of the IEEE
  International Conference on Control and Automation (ICCA)}, pp.~738--745,
  IEEE, 2019.

\bibitem{fox2003pddl2}
M.~Fox and D.~Long, ``Pddl2. 1: An extension to pddl for expressing temporal
  planning domains,'' {\em Journal of artificial intelligence research},
  vol.~20, pp.~61--124, 2003.

\bibitem{koenig2004design}
N.~Koenig and A.~Howard, ``Design and use paradigms for gazebo, an open-source
  multi-robot simulator,'' in {\em 2004 IEEE/RSJ International Conference on
  Intelligent Robots and Systems (IROS)(IEEE Cat. No. 04CH37566)}, vol.~3,
  pp.~2149--2154, IEEE, 2004.

\bibitem{wu2019detectron2}
Y.~Wu, A.~Kirillov, F.~Massa, W.-Y. Lo, and R.~Girshick, ``Detectron2.''
  https://github.com/facebookresearch/detectron2, 2019.

\bibitem{he2017mask}
K.~He, G.~Gkioxari, P.~Doll{\'a}r, and R.~Girshick, ``Mask r-cnn,'' in {\em
  Proceedings of the International Conference on Computer Vision (ICCV)},
  pp.~2961--2969, 2017.

\bibitem{cocoannotator}
J.~Brooks, ``{COCO Annotator}.'' https://github.com/jsbroks/coco-annotator/,
  2019.

\end{thebibliography}

\end{document}